\documentclass[letterpaper,journal]{IEEEtran}
\usepackage{amsmath,amsfonts}
\usepackage{amssymb} 
\usepackage{booktabs}
\usepackage{graphicx}
\usepackage{algorithm}
\usepackage{algpseudocode}
\usepackage{array}
\usepackage[caption=false,font=normalsize,labelfont=sf,textfont=sf]{subfig}
\usepackage{textcomp}
\usepackage{stfloats}
\usepackage{url}
\usepackage{verbatim}
\usepackage{graphicx}
\usepackage{cite}
\usepackage{tikz-cd}
\usepackage{tikz}
\usetikzlibrary{positioning, calc, arrows.meta}
\usepackage{fontawesome5}
\usepackage{hyperref}
\hypersetup{hidelinks}
\usepackage{xcolor}

\begin{document}

\title{Vector-Quantized World Model: Latent Feedback Motion Planning for Robot Manipulation}
\author{
Xuezhou Xu$^{*1}$,
Chung-Pang Wang$^{*1}$,
Zhenzhen Li$^{2}$,
and Benjamin Riviere$^{1}$%
\thanks{$^{*}$These authors contributed equally to this paper.}%
\thanks{$^{1}$New York University, New York, NY, USA. \quad $^{2}$NVIDIA.}%
}

% \author{IEEE Publication Technology,~\IEEEmembership{Staff,~IEEE,}
%         % <-this % stops a space
% \thanks{This paper was produced by the IEEE Publication Technology Group. They are in Piscataway, NJ.}% <-this % stops a space
% \thanks{Manuscript received April 19, 2026; revised June 12, 2026.}}

% % The paper headers
% \markboth{IEEE ROBOTICS AND AUTOMATION LETTERS,~Vol.~11, No.~6, June~2026}%
% {Shell \MakeLowercase{\textit{et al.}}: A Sample Article Using IEEEtran.cls for IEEE Journals}

% \IEEEpubid{0000--0000~\copyright~2026 IEEE}
% % Remember, if you use this you must call \IEEEpubidadjcol in the second
% % column for its text to clear the IEEEpubid mark.
\IEEEaftertitletext{%
    \vspace{-2em}
    {\centering
    \href{https://xuxuezhou.github.io/vq-wm/}
    {\faGlobe\enspace Website}
    \qquad
    \href{https://github.com/xuxuezhou/vqwm}
    {\faGithub\enspace Code}
    \par}
    \vspace{1em}
}
\maketitle

\begin{abstract}

Visual world models have shown strong potential for robot planning, yet long-horizon manipulation remains challenging due to compounding prediction errors, large search complexity, and latent plans that the low-level controller cannot reliably reach. In this work, we introduce Vector-Quantized World Model (VQ-WM), a latent feedback motion planning framework that integrates discrete long-horizon search with continuous visual feedback control: 
a discrete latent predictor models transitions between vector-quantized latent states, allowing the planner to look far ahead and find a shortest path to the goal through discrete search and a continuous learned controller sequentially reaches the planned states through visual feedback, correcting deviations to mitigate compounding errors. 
Whereas most vector quantization uses Euclidean distance, we use a novel estimator of the first-passage-time as a quasi-metric, which aligns the planning abstraction with the controller's capabilities. 
We evaluate on the physical robot manipulation of the Tower of Hanoi puzzle: whereas existing state-of-the-art foundation models (VLAs, WAMs) only succeed when the solution trace in their training dataset, our method achieves 100\% success rate with only short-play data, demonstrate its ability to compose long-horizon solutions beyond the short demonstrated trajectories and execute them reliably.
Our solution also outperforms strongest world-models by 8.0--85.3 percentage points on simulation benchmarks spanning manipulation and locomotion, demonstrating consistent progress and new capabilities. 
% that separates discrete planning from goal-conditioned feedback control.
% Using frozen pretrained visual features, Q-WM estimates first-passage time under the controller's closed-loop dynamics with a learned continuous latent predictor.
% These estimates identify states predicted to reach a common representative within a local control horizon, forming coverage regions that define the quantized planning states.
% A quantized latent predictor supports discrete search over these representatives, while the controller executes successive targets using visual feedback.
% \textcolor{red}{On physical Tower of Hanoi, Q-WM outperforms leading vision--language--action (VLA) and world action models (WAMs) from industry.}
% Simulation experiments further demonstrate strong performance on long-horizon manipulation tasks and thcompoundinge framework's applicability to navigation and locomotion.

\end{abstract}

\section{Introduction}
Visual world models combine latent representations with predictive dynamics, enabling planners to evaluate action sequences directly from visual inputs and select those that lead toward a desired long-term outcome before acting in the environment.
% Visual world models enable robots to learn predictive dynamics from image observations, reducing reliance on manually specified state representations.
% By representing desired outcomes as goal images, they support planning toward different objectives within a shared visual representation.
% \ben{what is the significance of visual here? I think we are missing something}. 
Recent advances in model architectures and training methods allow these models to learn from reward-free offline trajectories and perform zero-shot planning from visual observations~\cite{dino-wm,lewm}.
However, long-horizon manipulation tasks requiring multi-stage decision making and precise sequential execution remain challenging.
Existing world-model planners typically optimize action sequences through autoregressive rollouts in continuous latent space.
As the planning horizon grows, prediction errors can accumulate across rollout steps, while longer action sequences become harder to optimize.
These limitations constrain the effective planning horizon and can cause failure even when the dynamics model is accurate over short horizons.

% World models offer a promising approach to robot decision making by learning representations and predictive dynamics from observation--action trajectories~\cite{planet,visualforesight}.
% They enable a robot to evaluate candidate actions before executing them and to plan toward a goal specified by an image.
% Recent visual world models learn from reward-free offline data and support goal-conditioned planning in latent space~\cite{dino-wm,lewm}.
% This setting is particularly attractive for manipulation, where collected trajectories contain reusable interactions with objects and can support tasks specified after training.
% Long-horizon manipulation, however, requires these capabilities to extend across multiple stages, each combining a task-level decision with accurate local execution.

\begin{figure}[t]
    \centering
    \includegraphics[width=\columnwidth]{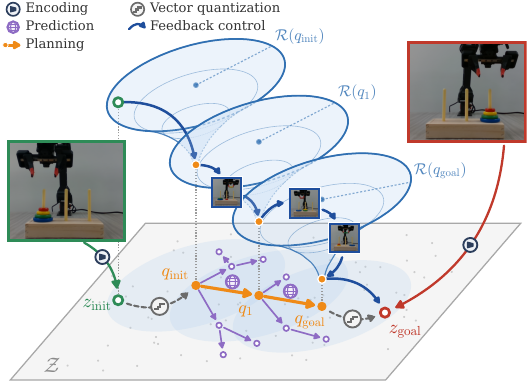}
    % \caption{Feedback motion planning in latent space.
    % Discrete planning connects quantized states, while local feedback control drives the system toward successive targets.
    % Region of attraction $\mathcal{R}(z^q)$ connect the continuous latent space $\mathcal{Z}$ to the planning abstraction.
    % \cpw{Overview of VQ-WM. The initial (green) and goal (red) images are encoded into latent states and vector-quantized based on controller reachability, enabling long-horizon reasoning over targets the controller is expected to reach. The planner performs discrete search over the quantized states using a discrete latent predictor. A continuous controller then sequentially reaches the planned targets using visual feedback for reliable long-horizon manipulation.}
    \caption{Overview of VQ-WM. The initial (green) and goal (red) images are encoded into latent states and vector-quantized according to controller reachability, so that each vector-quantized state serves as a target the controller is expected to reach. The planner performs discrete search over these vector-quantized states using a discrete latent predictor. A continuous controller then drives the robot toward each planned target in turn, using visual feedback to correct deviations along the way.}
    % \ben{[Done] label encoding, quantization, prediction, planning, feedback control. }
    % \ben{what is a "representative state"? what is a coverage region? be consistent with downstream notation, can we label quantization process and prediction process?}
    % \ben{make neural component bigger, and make it not an MLP, maybe some fancier architecture.}
    % \ben{Label clearly: Quantization (eg zinit to zq), Prediction (e.g. straight line from zq to zq'), and Feedback Control (e.g. curved line from zq to zq' that stabilizes error)
    % \ben{make funnels more transparent, make symbol font bigger, make quantized state topology a graph, not just a path}}
    % }
    \label{fig:motivation}
\end{figure}

% Recent work has improved long-horizon world-model planning by addressing compounding prediction errors and the difficulty of long-range action search.
% compounding prediction errors and the complexity of searching over long action sequences.
% Temporal hierarchy combines world models at different timescales to decompose long tasks into shorter planning problems~\cite{hwm}.
% Geometric regularization straightens latent trajectories to improve the planning objective~\cite{temporal}, while learned latent spaces at different abstraction levels support coarse-to-fine planning~\cite{hjepa}.

% Recent work has improved long-horizon world-model planning by addressing these limitations.
% Hierarchical world models decompose long tasks into shorter planning problems using models operating at different timescales~\cite{hwm}.
% Geometric regularization structures the latent planning space by straightening trajectories~\cite{temporal}.
% Other approaches learn multiple levels of latent abstraction for coarse-to-fine planning~\cite{hjepa}, and vector quantization has also been explored to construct hierarchical world models~\cite{director, thick, hilewm}.
Recent work has improved long-horizon world-model planning by addressing these limitations.
Hierarchical approaches such as H-JEPA learn multiple levels of latent abstraction to support coarse-to-fine planning~\cite{hjepa}.
Discrete latent representations have also been explored for hierarchical planning and control~\cite{director,thick}, while Hi-LeWM investigates vector-quantized macro-actions for high-level prediction and planning~\cite{hilewm}.
Hydra~\cite{hydra} uses vector-quantized multimodal intents for sampling-based navigation planning.
Other approaches improve the geometry of the latent planning space by encouraging locally straighter latent trajectories~\cite{temporal}.
% \ben{have people done hierarchical world models via vector quantization yet? if so, how are we different? if not, what are relative advantages of our approach?}
% \xuezhou{Hi-LeWM uses VQ to learn macro-action for a second-layer word model, but the paper is basically a simple combination of LeWM and HWM, also the performance is bad. Not sure if worth discussing.} 
% \ben{if they didnt put it on a robot we can just say first VQ on a robot}
% Despite these advances, planning and control errors can still reinforce each other: an inaccurate intermediate target can misdirect local control, while execution errors can invalidate subsequent planning stages.
Despite these advances, plans formed in latent space may still be difficult to execute reliably.
Replanning can correct deviations, but does not by itself ensure that proposed targets are compatible with the controller's capabilities.
The planning abstraction should reflect which states the controller can reliably reach and stabilize, a principle central to classical \emph{feedback motion planning}~\cite{sequential_composition,funnels}.
% This motivates a planning abstraction that reflects the controller's ability to reach intermediate targets and correct local deviations.

% To incorporate controller capabilities into the planning structure, we revisit classical \emph{feedback motion planning}~\cite{sequential_composition,funnels}.
% LQR-Trees proposes an approach that organizes trajectories using local feedback controllers and their regions of attraction~\cite{lqrtrees}.
% Classical \emph{feedback motion planning} addresses this by incorporating the controller's stabilization capabilities into the planning process~\cite{sequential_composition,funnels}.
% LQR-Trees constructs a tree of trajectories paired with local feedback controllers and their regions of attraction~\cite{lqrtrees}.
% Each region characterizes a set of initial conditions from which feedback control can stabilize the associated motion, and these regions organize how the system can progress toward the goal.
% The planning structure therefore incorporates the controller's ability to correct local deviations.
% This perspective suggests that a useful planning unit is a target together with the states from which local feedback can reach it.
The LQR-Trees framework makes this connection explicit by associating each trajectory with a feedback controller and a region of attraction, characterizing the states from which the controller can recover deviations and steer the system toward the goal~\cite{lqrtrees}.
Motivated by these limitations and the principle of coupling planning and control, we propose \emph{Vector-Quantized World Model} (VQ-WM), a \emph{latent feedback motion planning} framework for long-horizon robotic manipulation.
As illustrated in Fig.~\ref{fig:motivation}, VQ-WM uses vector quantization to construct a discrete planning space whose vector-quantized states serve as targets for visual feedback control.
The planner runs discrete search on the resulting graph to find a path to the goal, and the controller runs tracking control between consecutive vector-quantized states on the path. 
% Whereas standard vector quantization uses Euclidean distance to compute centroids, we use a controller-aware distance, which certifies that each planning step is executable by the downstream controller. 
The central design choice is to define vector quantization through the controller’s ability to reach these targets, aligning the planning abstraction with the controller’s capabilities.
In particular, we use the first-passage time, which is the number of controller calls until a trajectory driven by the controller first enters a neighborhood of its target. 
% We propose a novel way to estimate this distance from offline data by simulating the controller in closed loop inside a learned world model.
We estimate this quantity through closed-loop controller rollouts in a latent world model learned from offline data, and use it to construct the state codebook and assign continuous latent states to vector-quantized states.

Our main contributions are:
\begin{itemize}
    \item We introduce VQ-WM, a latent feedback motion planning framework that uses vector quantization to integrate discrete long-horizon search with continuous visual feedback control.
    % \item We introduce VQ-WM, a latent feedback motion planning framework that uses vector quantization to connect \ben{integrate?, interweave?} discrete long-horizon search with continuous visual feedback control. 
    % \ben{We introduce Q-WM, a latent feedback motion planning framework that uses vector quantization to automatically integrate discrete long-horizon search and continuous control.}
    % \ben{We introduce Q-WM, that uses vector quantization to enable latent feedback motion planning framework couple discrete long-horizon search with continuous feedback control.}
    % \item We use first-passage time estimated through closed-loop controller rollouts in a learned latent world model to define vector state quantization, aligning the planning abstraction with the controller’s capabilities.
    \item We propose first-passage time as a measure of controller reachability for state vector quantization and estimate it through closed-loop controller rollouts in a learned latent world model. The first-passage time from one state to another is the number of controller calls until the closed-loop trajectory first reaches it. This aligns the planning abstraction with the controller’s capabilities and supports effective planning with compact state codebooks.
    % \ben{We propose the first-passage time as a quasi-metric for vector quantization; the first-passage time between two states is the expected number of timesteps between the two states under the current policy, and its effect on vector quantization is that it groups states that are reachable by the controller, aligning the planning abstraction with the controller’s capabilities. 
    % We propose an new estimation of first-passage time using our world models. }
    % \ben{We propose the first-passage time under closed-loop dynamics as a distance for vector quantization. This is conceptually similar to planning over a controller's reachable set and coupled planning and control in a natural way.}
    % Through extensive experiments, we validate Q-WM’s long-horizon reasoning and robust execution capabilities, enabled by our vector quantization and distance function mechanisms......
    % We validate VQ-WM through hardware and simulation experiments. Our controller-aware state quantization substantially improves long-horizon manipulation performance, with VQ-WM achieving nearly 100\% task progress on physical four-disk Tower of Hanoi and outperforming the strongest baselines by 13.3–28.0 percentage points on challenging simulated manipulation tasks. Decoupling planning and control further allows the controller to incorporate proprioceptive information without modifying the planner, increasing Walker success from 22.0\% to 97.3\%.
    \item We validate VQ-WM in hardware and simulation. Learning from short play data without full-task demonstrations, it solves the physical Tower of Hanoi puzzle with 100\% task progress, outperforming the competitive industrial baselines by at least 95 percentage points. In simulation, it surpasses the strongest world-model planner, H-JEPA, by 13.3--34.6 percentage points on challenging long-horizon manipulation tasks, and adding proprioceptive inputs to the controller raises Walker success from 22.0\% to 97.3\%.
% \ben{add numbers and be more specific about the mechanisms that enable better performance.}
\end{itemize}

% \zl{
% Our main contributions are:}

% \begin{itemize}
%     \item \zl{\textbf{Controller-aware state abstraction.} We introduce a first-passage-time-based quantization method that uses closed-loop rollouts in a learned world model to construct planning states informed by the reachability of a learned controller.}
%     \item \zl{\textbf{Long-horizon latent feedback planning.} We develop Q-WM, which combines discrete search over quantized states with continuous visual feedback control, enabling the composition of local transitions into long-horizon plans.}
%     \item \zl{\textbf{Real-world and simulation validation.} We demonstrate Q-WM's ability to compose short play trajectories into full Tower of Hanoi solutions without full-task demonstrations, achieve strong performance across simulated manipulation benchmarks, and maintain high success with compact state codebooks through controller-aware quantization.}
% \end{itemize}

% \ben{first contribution is vector quantization for latent feedback motion planning, second contribution is selection of distance function as mean first passage time to couple planning and control, third contribution is validating both of these mechanisms in hardware and simulation, and solving the Towers of Hanoi problem.}

\section{Related Work}
% \ben{I think this section needs a little bit of work: first paragraph's main point is that existing things dont constuct discrete planning states based on reachability. Second paragraphs point is also that WM needs discrete states based on reachability. Third paragraphs point is missing what we are doing. I propose: first paragraph is all "single layer WMs", and fails becuae optimization of long horizons or something is hard. second paragraphs is "quantization is good, but needs to be coupled to reachability. third paragrpah needs better intro sentence, and better conclusion sentence: what is the definitiion of feedback motion planning? are we the first feedback motion planning WM?}
% \cpw{we are not the first first feedback motion planning WM. (Check: Pixels to Proofs: Probabilistically-Safe Control in Latent World Models via Parallel Conformal Robust MPC~\cite{sls2_2026}}

\subsection{World Models for Visual Planning}
World models learn predictive dynamics from observations to support action selection.
Early approaches combine visual representations with latent dynamics for control~\cite{e2c,planet}, or predict future images for goal-directed manipulation~\cite{visualforesight}.
Universal Planning Networks jointly learn representations and differentiable action planning~\cite{upn}, while Dreamer learns behavior through latent imagination~\cite{dreamer} and TD-MPC methods combine short-horizon latent planning with value estimation~\cite{tdmpc,tdmpc2}.
More recent methods support image-goal planning from pretrained visual features or reward-free offline trajectories~\cite{dino-wm,lewm,pldm,vjepa2}.
% Fast-LeWM further improves planning efficiency by predicting future latent states in parallel from action prefixes~\cite{fast-lewm}. 
% Planning efficiency is improved through alternative rollout mechanisms~\cite{fast-lewm}.
However, continuous action-sequence planning becomes increasingly challenging over long horizons as search complexity grows rapidly and prediction errors can accumulate during autoregressive rollouts. 
VQ-WM instead performs discrete search over vector-quantized states without optimizing the full sequence of low-level actions.

% Planning efficiency is improved through alternative rollout mechanisms~\cite{fast-lewm}.
% However, these methods do not explicitly construct discrete planning states based on reachability under a feedback controller.
% This distinction matters because proximity in a learned latent space does not imply that the controller can reach the corresponding planning target.
% VQ-WM constructs discrete states according to the controller's ability to reach their vector-quantized states, and learns transitions for discrete search.

% \ben{However, none of these works do XX, which is important for YY.}
% Q-WM focuses on the planning abstraction: it constructs discrete states using the behavior of a separately trained controller and learns quantized transitions for discrete search.

\subsection{Abstractions for Long-Horizon Planning}
Temporal hierarchies decompose extended tasks into shorter planning problems ~\cite{director}. 
HWM learns world models at multiple temporal scales and demonstrates hierarchical planning for real-world pick-and-place from a single goal image ~\cite{hwm}.
H-JEPA combines multiple prediction horizons with latent spaces at different abstraction levels~\cite{hjepa}.
However, predicted subgoals can still be difficult for the execution mechanism to realize~\cite{hilewm}.
% Temporal Straightening shapes latent geometry to improve planning~\cite{temporal} \ben{does this reference use hierarchies and does it belong here?}.
Vector quantization provides a compact state space for efficient discrete search, but the resulting state assignments must also reflect controller reachability. 
Graph-based methods incorporate reachability into planning through learned goal-conditioned control. 
SoRB searches over replay-buffer states using distances learned through goal-conditioned reinforcement learning~\cite{sorb}, while L3P learns latent landmarks and graph edges from reachability estimates distilled from value functions~\cite{l3p}.
VQ-WM vector-quantizes states based on first-passage times estimated through closed-loop controller rollouts in a learned world model, then learns vector-quantized transitions from offline trajectories for discrete search. 
This couples the planning abstraction to the controller’s ability to reach the assigned targets.

% Graph-based methods also connect planning to learned goal-conditioned control.
% SoRB searches over replay-buffer states using distances learned through goal-conditioned reinforcement learning~\cite{sorb}, while L3P learns latent landmarks and graph edges from reachability estimates distilled from value functions~\cite{l3p}.
% \ben{However, none of these works do XX, which is important for YY.}
% However, these approaches do not define state vector quantization through explicit controller rollouts, which is important for aligning discrete state assignments with the controller's ability to reach their vector-quantized states.
% VQ-WM instead constructs vector quantization coverage sets from first-passage time estimated by rolling out a fixed controller through a continuous latent predictor, then learns a vector-quantized transition model from offline trajectories.
% \ben{has anyone put temporal abstracted wm on hardware?}

\subsection{Feedback Motion Planning and Goal-Conditioned Control}
Feedback motion planning accounts for feedback control during planning to avoid trajectories that are difficult or costly to stabilize~\cite{lqrtrees}.
% Planning without accounting for feedback can produce trajectories that are difficult or costly to stabilize~\cite{lqrtrees}.
Classical approaches compose local controllers so that each brings the system into a region where the next can succeed, using domains of attraction or trajectory funnels to characterize these regions~\cite{sequential_composition,lqrtrees,funnels}.
% Classical feedback motion planning addresses this challenge by composing local controllers so that each brings the system into a region where the next can succeed, using domains of attraction or trajectory funnels to characterize these regions~\cite{sequential_composition,lqrtrees,funnels}.
% Classical feedback motion planning connects local controllers through their domains of attraction~\cite{sequential_composition,lqrtrees,funnels}.
% Goal-conditioned methods learn reusable behaviors from play~\cite{play_lmp} or generate action sequences through inverse dynamics~\cite{glamor}.
Goal-conditioned methods learn reusable behaviors from play~\cite{play_lmp} or generate goal-directed action sequences through inverse dynamics models~\cite{glamor}.
Composing such behaviors into long-horizon plans requires accounting for the states from which each target can be reached.
VQ-WM applies this composition principle to a learned goal-conditioned controller in visual latent space. 
Instead of formally verifying funnels, it estimates controller reachability using first-passage times from closed-loop rollouts in a learned world model. 
These estimates guide the composition of local behaviors into long-horizon plans, which the controller executes using continuous visual feedback.
We demonstrate reliable long-horizon manipulation on real hardware using this latent feedback motion planning framework.

\section{Problem Definition and Background}\label{sec:problem}
% \cpw{structures of this section: dataset paragraph and fpt, Model components list for us to keep track of: }

% \xuezhou{define the concept (not estimation formula) for FPT, and quantizer notation $\phi_Z(z)$ and $\hat{\phi}_Z(z)$ (quantize $z_{goal}$ needs the reverse) here.}

% \xuezhou{I have merged FPT estimation in State Vector Quantization in training.}

% \xuezhou{To connect discrete planning with continuous control, the state vector quantizer $\phi_Z$ should reflect which latent targets the learned controller $\pi$ can reach. 
% We characterize this reachability using first-passage time, which measures the number of controller calls required for a closed-loop trajectory to first enter a neighborhood of its target. 
% By capturing both whether and how quickly the controller reaches a target, this quantity provides a controller-dependent basis for state vector quantization.
% The first-passage time from $z$ to $z'$ is the number of controller calls until the closed-loop trajectory first enters an $\epsilon$-ball around $z'$, where $\epsilon>0$ is the arrival tolerance.}

% \zl{Geometric proximity in latent space does not necessarily reflect the effort required by a feedback controller to reach a target. We therefore use first-passage time as a directed, controller-dependent reachability cost.} 

Our work follows a vision-based control framework, which models the environment as a controlled Markov process $\langle \mathcal O,\mathcal A,P_O\rangle$, where $\mathcal O$ is the observation space and $\mathcal A$ is the physical action space. 
At each time step $t$, the robot receives an observation $o_t\in\mathcal O$ and applies an action $a_t\in\mathcal A$. 
The transition distribution $P_O(o_{t+1}\mid o_t,a_t)$ describes how the robot’s observation changes in response to its action.

Given a pre-collected, reward-free offline dataset $\mathcal D=\{(o_0^{(n)},a_0^{(n)},\ldots,a_{T_n-1}^{(n)},o_{T_n}^{(n)})\}_{n=1}^{N}$ of $N$ observation–action trajectories with respective episode lengths $T_n$, our goal at test time is to steer the system from an initial observation $o_{\mathrm{init}}$ to a state specified by a goal image $o_{\mathrm{goal}}$.

\section{Methods}

% \zl{At a high level, VQ-WM constructs a controller-aware discrete planning space and adapt it with a feedback controller. During offline learning, we train a goal-conditioned controller and a continuous latent predictor, then use their closed-loop rollouts to estimate first-passage times for state quantization. We subsequently learn a discrete transition model over quantized states. At inference time, graph search selects a sequence of intermediate targets, which the controller tracks using visual feedback.
% We can add a high level walk through of the mehtod design, and also include the definition of "quantize" - vector quantize. Or just simply replace all quantize as vector quantize.
% }

% \section{Methods}
\label{sec:methods}

% \todo{
% Change method organization to: 
% \begin{itemize}
%     \item Make a supporting figure that shows all the components of our architecture 
%     \item Split methods into subsections: short overview paragraph (notation and components), inference, training, and mean-first passage time estimation 
%     \item Change Controller-Aware distance to mean-first passage time. This object exists, we should call it by its proper name 
%     \item current organization is not good 
% \end{itemize}
% }
% \cpw{Strutures of this section: overview, inference, training, estimation)}

\subsection{Overview}
\label{sec:overview_}
% \zl{
% % At a high level, VQ-WM constructs a controller-aware discrete planning space and adapt it with a feedback controller. 
% % During offline learning, we train a goal-conditioned controller and a continuous latent predictor, then use their closed-loop rollouts to estimate first-passage times for state quantization. 
% % We subsequently learn a discrete transition model over quantized states. 
% % At inference time, graph search selects a sequence of intermediate targets, which the controller tracks using visual feedback.
% [Done] We can add a high level walk through of the method design, and also include the definition of "quantize" - vector quantize. 
% Or just simply replace all quantize as vector quantize.
% }

At a high level, VQ-WM interweaves discrete planning over a controller-aware state codebook $\mathcal Q$ with continuous feedback control using $\pi$.

The components of VQ-WM are $\{E_Z,\allowbreak E_U,\allowbreak P_Z,\allowbreak P_Q,\allowbreak \phi_Z,\allowbreak \phi_U,\allowbreak \pi\}$. 
% The pretrained, frozen encoder $E_Z$ maps observations to continuous latent states $z_t=E_Z(o_t)\in\mathcal Z$, 
% and we use V-JEPA~2~\cite{vjepa2} in this work, while $E_U$ maps pairs of vector-quantized states to continuous latent actions. 
The pretrained, frozen encoder $E_Z$ maps observations to continuous latent states $z_t=E_Z(o_t)\in\mathcal Z$. For example, in this work, we use V-JEPA~2~\cite{vjepa2}. 
The trained encoder $E_U$ maps pairs of vector-quantized states to continuous latent actions. 
The quantizers $\phi_Z$ and $\phi_U$ assign these representations to entries in finite state and action codebooks $\mathcal Q\subset\mathcal Z$ and $\mathcal U$, respectively. 
The predictors $P_Z$ and $P_Q$ model continuous and vector-quantized latent transitions, respectively, and the goal-conditioned controller $\pi(\cdot \mid z, z')$ maps the current and target latent states to a physical action.

A natural choice for the state quantizer $\phi_Z$ is to assign each latent state to its nearest codebook entry under Euclidean distance. 
However, geometric proximity in latent space does not necessarily reflect the effort required by a feedback controller to reach a target. 
We therefore use \textit{first-passage time} (FPT) as a directed, controller-dependent reachability cost for state vector quantization. 
FPT measures the number of controller calls required for a closed-loop trajectory to first enter a neighborhood of its target. 
Using such a distance measure for state vector quantization couples discrete planning with continuous control, enabling efficient search over targets aligned with the controller’s capabilities.

\begin{figure*}[t]
    \centering
    \includegraphics[width=\linewidth]{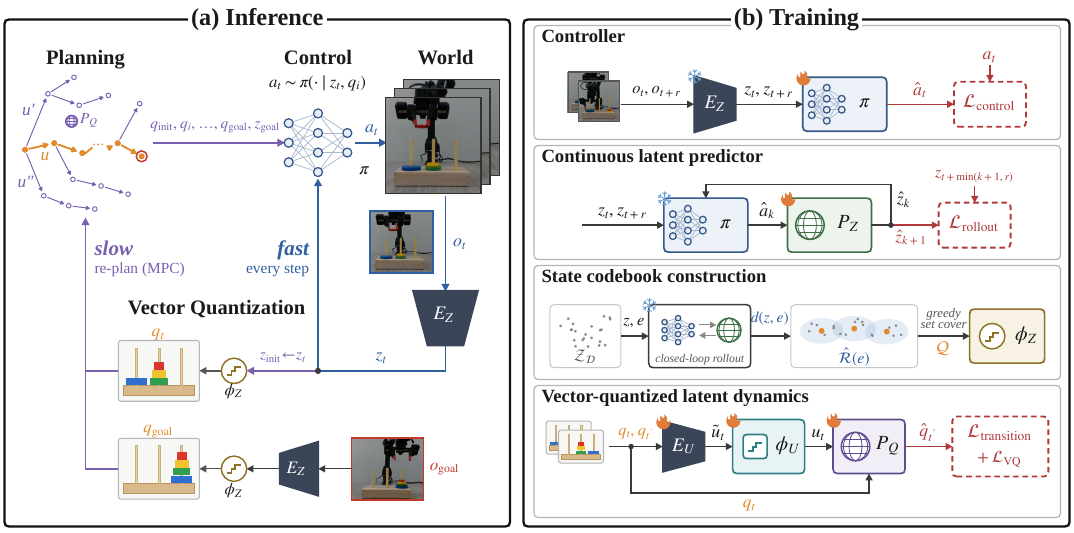}
    \caption{VQ-WM components and pipeline.
    (a) Inference encodes the initial and goal images, searches over state codes, and executes the planned targets with visual feedback. 
    % \ben{it is not obvious where the "real" robot is in the inference plot, right now it looks like it is part of feedback control. I think $o_{init}$ should be an instance of $o_t$.} 
    % \ben{can we put "vector quantization" above the graphic that describes it? to be consistent with rest of figure.}
    (b) Offline training learns the controller and continuous latent predictor, performs state vector quantization using first-passage time, and then learns vector-quantized latent dynamics.
    % \ben{better use of "frozen" and "fire" emoji. only show these in training, and always show fire if you show frozen}
    % \ben{I am not sure this is the best inference diagram: I find it nicer to draw analogy to classical feedback, like the diagram I put in the methods section. Also, this looks AI generated, and I think people will respond poorly to that. }
 }
    \label{fig:qwm_overview}
\end{figure*}

During offline learning, we first train $\pi$ and the continuous latent predictor $P_Z$, which predicts the next latent state as $\hat{z}_{t+1}\sim P_Z(\,\cdot\mid z_t,a_t)$.
Their closed-loop rollouts provide first-passage time estimates $d(z,z')$, which we use to construct $\mathcal Q$ and quantize continuous latent states $z\in\mathcal Z$ as $q=\phi_Z(z)\in\mathcal Q$.
With $\mathcal Q$ fixed, we jointly learn the latent action encoder $E_U$, action codebook $\mathcal U$, and discrete latent predictor $P_Q$.
The predictor $P_Q$ outputs a categorical distribution $P_Q(q'\mid q,u)$ over the next vector-quantized state, conditioned on the current vector-quantized state $q$ and vector-quantized action $u$.
At inference time, our planner performs graph search using $P_Q$ to select a sequence of intermediate targets in $\mathcal Q$, which $\pi$ tracks using visual feedback.
The inference and training pipelines of VQ-WM are shown in Fig.~\ref{fig:qwm_overview}.

\subsection{Inference}
\label{sec:inference}

% \ben{swap fig 2b and 2a}
% As shown in Fig.~\ref{fig:qwm_overview}(a) and Algorithm~\ref{alg:inference}, inference alternates between planning and control.
VQ-WM performs discrete planning and continuous feedback control at different rates, as shown in Fig.~\ref{fig:qwm_overview}(a).
Planning operates at a slower rate and uses the discrete latent predictor $P_Q$ to expand the search graph online, while the goal-conditioned controller $\pi$ operates at a faster rate with visual feedback.

% Inference process alternates between planning and control with different rate.
% Planning operates at a slower rate by expanding rollouts of the discrete latent predictor $P_Q$ online, while control operates at a faster rate by repeatedly querying $\pi$ with visual feedback.

\subsubsection{Planning Loop}
Our planner searches the vector-quantized state graph for a shortest path to the goal. 
Given initial and goal observations, we encode them as $z_{\mathrm{init}}=E_Z(o_{\mathrm{init}})$ and $z_{\mathrm{goal}}=E_Z(o_{\mathrm{goal}})$, then quantize them into the vector-quantized states $q_{\mathrm{init}}$ and $q_{\mathrm{goal}}$.

% Planning \cpw{Our planner} searches on the vector-quantized state graph for a path to the goal.
% We encode the initial and goal images as $z_{\mathrm{init}}=E_Z(o_{\mathrm{init}})$ and $z_{\mathrm{goal}}=E_Z(o_{\mathrm{goal}})$, then vector quantize them into corresponding vector-quantized states $q_{\mathrm{init}}=\phi_Z(z_{\mathrm{init}})$ and
% $q_{\mathrm{goal}}=\bar{\phi}_Z(z_{\mathrm{goal}})$.
Starting from $q_{\mathrm{init}}$, our planner explores successor states predicted by $P_Q(\cdot\mid q,u)$ for all vector-quantized actions $u\in\mathcal U$ at each expanded state $q$.
It builds the graph online by continuing to expand newly discovered vector-quantized states until $q_{\mathrm{goal}}$ is found.
The predicted categorical distributions define a stochastic transition graph that can be used with stochastic shortest-path algorithms.
In our implementation, we retain the most likely successor $\arg\max_{q'\in\mathcal Q}P_Q(q'\mid q,u)$ for each state–action pair and use breadth-first search (BFS) to find a shortest path from $q_{\mathrm{init}}$ to $q_{\mathrm{goal}}$ with the fewest graph transitions.
We denote this path by the sequence of vector-quantized states $(q_0,q_1,\ldots,q_L)$, where $q_0=q_{\mathrm{init}}$ and $q_L=q_{\mathrm{goal}}$. 
If the goal is predicted to be reachable within $R$ controller calls, $d(z_{\mathrm{init}},z_{\mathrm{goal}})\leq R$, we bypass graph search and use $\pi$ with $z_{\mathrm{goal}}$ as its target.

% Our planner can operate in a receding-horizon manner by periodically updating the plan from the current observation and executing only its initial portion before replanning.
% In our implementation, the efficiency of our planner allows us to compute a complete plan at the start of each episode.
% To evaluate the stability of our feedback motion planning framework, we use an infinite replanning interval to plan once per episode and execute the resulting plan using continuous visual feedback.

Our planner can be implemented in a receding-horizon manner by periodically updating the plan from the current observation and executing only its initial portion before replanning, or in open-loop by setting the replanning frequency to zero. 
The quantization procedure makes planning computationally efficient, allowing VQ-WM to compute a complete plan spanning approximately five minutes of robot execution in under one second on an NVIDIA RTX 5080 GPU.
% The quantization procedure makes the computational cost of planning cheap, we run at \ben{XX}Hz on \ben{XX} hardware. 
In our experiments, to evaluate the stability of our feedback motion planning framework, we use open-loop planning once per episode and execute the resulting plan using continuous visual feedback.

\subsubsection{Control Loop}
The continuous controller operates at a higher rate than the planner, tracking the planned path using visual feedback.
At each control step, we encode the current observation as $z_t=E_Z(o_t)$ and apply an action $\hat{a}_t\sim\pi(\cdot\mid z_t,q_i)$ toward the current planned state $q_i$.
The controller follows the planned states in order, beginning with $q_{\mathrm{init}}$, and finally target $z_{\mathrm{goal}}$.
For each planned transition, it runs for the number of controller calls specified by the first-passage time estimated between consecutive planned states before switching to the next target.
Execution succeeds when the task is completed and fails if it remains incomplete after the final target has been attempted.

\subsection{Training}
\label{sec:training}

% We learn Q-WM from an offline dataset $\mathcal{D}=\{\tau_n\}_{n=1}^{N}$ of observation--action trajectories. \ben{no need to introduce new notation tau, we used o,a before, be consistent}
% \ben{We train the components of $Q-WM = \{E_\theta...\}$ from an offline dataset of observation action trajectories: $D = \{(o_t,a_t)\}$. This section is visualized in Fig.~\ref{fig:qwm_overview}(a). . The encoder, $E_\theta$ is ...}
We train VQ-WM from the offline dataset $\mathcal{D}$ defined in Sec.~\ref{sec:problem}, following the pipeline in Fig.~\ref{fig:qwm_overview}(b).
We first train the goal-conditioned controller $\pi$ and continuous latent predictor $P_Z$, then construct the state quantizer $\phi_Z$ and codebook $\mathcal Q$, and finally learn the vector-quantized latent dynamics.
A pretrained latent state encoder $E_Z$ maps observations to latent states $z_t=E_Z(o_t)$ and remains frozen throughout training.

% We encode observations as $z_t=E_\theta(o_t)$ using a pretrained visual encoder (e.g., V-JEPA~2~\cite{vjepa2}), whose parameters remain frozen throughout training.
% As shown in Fig.~\ref{fig:qwm_overview}(b), we first train $\mathrm{ID}_A$.
% We then train $P_{Z^c}$ with $\mathrm{ID}_A$ frozen, making the predictor controller-aware by learning the closed-loop dynamics induced by the trained controller.
% Their closed-loop rollouts are used to construct the state codebook (Sec.~\ref{sec:fpt}).
% With the state codes fixed, we jointly train $\mathrm{ID}_U$, the action codebook, and $P_{Z^q}$.
% We first train the controller $\pi$, then freeze it when training the continuous latent predictor $P_{Z}$.
% This makes $P_{Z}$ controller-aware by learning the closed-loop dynamics induced by $\pi$.
% We use rollouts of these two components to construct the state codebook.
% With the state codebook fixed, we jointly train $E_U$, the action codebook, and $P_{Q}$.
% \ben{this is too much signposting. I think we can remove this paragraph.}

\subsubsection{Controller}

We train the goal-conditioned controller $\pi$ on state–goal pairs $(z_t,z_{t+r})$ from the same offline trajectory, sampling $r\sim\mathrm{Uniform}\{0,\dots,R\}$ to provide targets at different temporal distances.
When $r=0$, we replace $a_t$ with a hold action so the controller learns to maintain the current state.
We train $\pi$ by minimizing:
\begin{equation}
  \mathcal{L}_{\mathrm{control}}
  = \big\|a_t-\hat a_t\big\|_1,
  \quad
  \hat a_t
  = \mathbb{E}_{a\sim\pi(\cdot\mid z_t,z_{t+r})}[a].
  \label{eq:control}
\end{equation}
For a target $z'$, we approximate its $R$-step region of attraction $\mathcal R(z')$ as the set of initial latent states from which closed-loop control with $\pi$ reaches an $\epsilon$-neighborhood of $z'$ within $R$ controller calls with arrival tolerance $\epsilon>0$.

% We train $\pi$ with  $r \sim \mathrm{Uniform}\{0,\dots,R\}$.
% We sample state--target pairs $(z_t,z_{t+r})$ from the same trajectory with $r \sim \mathrm{Uniform}\{0,\dots,R\}$ and supervise the first action toward the target \ben{[Done] I dont understand the point of randomly sampling "r"... isnt the first action after the current state $z_t$ always the same regardless of $z_{t+r}$?}:
% \xuezhou{It may not be true when multiple trajectories go through one same state. Simple Example: Let's say two trajectories with only two states both start with $z_0$, end with $z_1$ and $z_2$ respectively. The action will condition on the second input. Same case applied to longer trajecoties.}
% \begin{equation}
%   \mathcal{L}_{\mathrm{control}} = \big\| a_t - \pi(z_t, z_{t+r}) \big\|_1 .
%   \label{eq:ida_loss}
% \end{equation}
% \begin{equation}
%   \mathcal{L}_{\mathrm{control}} = \big\| a_t - \hat{a}_t \big\|_1, \quad
%   \hat{a}_t = \mathbb{E}_{a \sim \pi(\cdot \mid z_t, z_{t+r})}[a].
%   \label{eq:control}
% \end{equation}

% When $r=0$, $\pi$ outputs a hold action that maintains the current state; this action also replaces $a_t$ as the supervision when $r=0$.
% For a target $z'$, $\mathcal{R}(z')$ denotes its $R$-step region of attraction: the initial latent states from which closed-loop control with $\pi$ reaches its $\epsilon$-neighborhood within $R$ calls~\cite{lqrtrees}.

\subsubsection{Continuous Latent Predictor}
The continuous latent predictor $P_Z$ predicts the next latent state from the current latent state and a physical action.
We supervise its predictions over closed-loop rollouts with $\pi$, matching the rollout procedure used for first-passage time estimation.
% The continuous latent predictor $P_Z$ learns to predict the next latent state from the current latent state and an action.
% We supervise closed-loop rollouts to match its use for first-passage time estimation.
Given a latent state $z_t$ and a target $z' = z_{t+r}$ with $r \sim \mathrm{Uniform}\{0, \dots, R\}$, we initialize $\hat z_0 = z_t$ and predict $R$ steps with the closed-loop rollout.
The predicted trajectory is supervised with the recorded one, which is held at the target once it is reached:
\begin{equation}
  \mathcal{L}_{\mathrm{rollout}}
  = \frac{1}{R} \sum_{k=0}^{R-1}
  \big\| \hat z_{k+1} - z_{t+\min\{k+1,\, r\}} \big\|_2^2 .
  \label{eq:pred_loss}
\end{equation}

\begin{algorithm}[t]
\caption{Codebook Construction via Greedy Set Cover}
\label{alg:setcover}
\small
\begin{algorithmic}[1]
\Require Latent states $\mathcal{Z}_{\mathcal D}$,
first-passage time estimate $d$,
controller-call budget $R$, number of neighbors $K$
\Ensure State codebook $\mathcal{Q}$,
vector-quantized states $\{q\}_{z \in \mathcal{Z}_{\mathcal D}}$

\For{each $e \in \mathcal{Z}_{\mathcal D}$}
    \Comment{approximate regions of attraction}
    \State $\hat{\mathcal{R}}(e) \gets \{e\}$
    \For{each $z_{n,t} \in \mathrm{KNN}(e,K)$
    with $\rho = d(z_{n,t},e) \le R$}
        \State $\hat{\mathcal{R}}(e)
        \gets \hat{\mathcal{R}}(e) \cup
        \{z_{n,t-m}\}_{m=0}^{\min(R-\rho,t)}$
    \EndFor
\EndFor

\State $\mathcal{Q} \gets \emptyset$,
$\mathcal{V} \gets \mathcal{Z}_{\mathcal D}$
\Comment{greedy set cover}
\While{$\mathcal{V} \neq \emptyset$}
    \State $e^\star \gets
    \arg\max_{e \in \mathcal{Z}_{\mathcal D}}
    |\hat{\mathcal{R}}(e) \cap \mathcal{V}|$
    \State $\mathcal{Q} \gets \mathcal{Q} \cup \{e^\star\}$,
    $\mathcal{V} \gets
    \mathcal{V} \setminus \hat{\mathcal{R}}(e^\star)$
\EndWhile

\State assign each $z \in \mathcal{Z}_{\mathcal D}$ to
$q = \phi_Z(z)$
\State \Return $\mathcal{Q}$,
$\{q\}_{z \in \mathcal{Z}_{\mathcal D}}$

\end{algorithmic}
\end{algorithm}

\subsubsection{State Vector Quantization}
% We first estimate first-passage time through closed-loop rollouts of $\pi$ using the continuous latent predictor $P_Z$:
We estimate the first-passage time from $z$ to $z'$ by rolling out the controller $\pi$ toward $z'$ in closed loop with the continuous latent predictor $P_Z$: 
\begin{equation}
\begin{aligned}
  &\hat{a}_k \sim \pi(\cdot \mid \hat{z}_k, z'), \quad
   \hat{z}_{k+1} \sim P_Z\big(\cdot \mid \hat{z}_k, \hat{a}_k\big), \quad
   \hat{z}_0 = z, \\
  &d(z, z') = \min\big\{ k \in \{0, \dots, R\} : \|\hat{z}_k - z'\|_2 \le \epsilon \big\}.
\end{aligned}
\label{eq:fpt}
\end{equation}
We set $d(z,z')=\infty$ if the target is not reached within $R$ calls.
This first-passage time estimate guides construction of the state codebook.
% Here, $\epsilon>0$ is the latent-space arrival tolerance and $R$ is the maximum number of controller calls, and we set $d(z,z')=\infty$ if the target is not reached within this budget.
% The estimate is directed, since reaching $z'$ from $z$ does not imply the reverse. 
% This first-passage time estimate guides construction of the state codebook. 
During inference, it is used to vector quantize the initial and goal states and determine the number of controller calls allocated to each planned target.

Let $\mathcal{Z}_{\mathcal D} \subseteq \mathcal{Z}$ denote the latent states encoded from the dataset, with $z_t^{(n)}=E_Z(o_t^{(n)})$ for step $t$ of trajectory $n$.
For each candidate code $e\in\mathcal{Z}_{\mathcal D}$, we identify states $z$ from which $\pi$ can reach $e$ within $R$ calls, using $d(z,e)$:
\begin{equation}
  \hat{\mathcal{R}}(e)=\{z\in\mathcal{Z}_{\mathcal D}:d(z,e)\leq R\}.
  \label{eq:coverage}
\end{equation}
This set estimates the $R$-step region of attraction of $e$.
We seek the smallest codebook whose estimated regions of attraction collectively cover $\mathcal{Z}_{\mathcal D}$, and use greedy set cover to approximate this selection (Fig.~\ref{fig:qwm_overview}(b)).

For efficient computation, we approximate $\hat{\mathcal{R}}(e)$ by using local rollout evaluations and recorded trajectories.
Starting with all states uncovered, greedy set cover repeatedly adds the code covering the largest number of uncovered states and marks these states as covered.
The procedure stops when the selected sets cover the entire encoded dataset, $\mathcal{Z}_{\mathcal D}
  =\bigcup_{q\in \mathcal{Q}}\hat{\mathcal{R}}(q)$.
The vector quantizer then assigns each state to the code it reaches in the fewest controller calls: 
% and its reverse assigns a state to the code from which it is reached in the fewest calls,
\begin{equation}
\phi_Z(z)=\arg\min_{q\in\mathcal{Q}} d(z,q).
% \qquad
% \bar{\phi}_Z(z)=\arg\min_{q\in\mathcal{Q}} d(q,z).
\label{eq:quantizer}
\end{equation}
Since $d$ is directed, we assign the goal $z_\mathrm{goal}$ to the code from which the controller can reach it in the fewest calls, $q_{goal}=\arg\min_{q\in\mathcal{Q}} d(q, z_{goal})$.
Algorithm~\ref{alg:setcover} details region approximation, greedy selection, and state assignment.

\subsubsection{Vector-Quantized Latent Dynamics}
With the state codebook fixed, we jointly learn the latent action encoder, the action codebook and the discrete latent predictor (Fig.~\ref{fig:qwm_overview}(b)).
The fixed state codebook $\mathcal Q$ consists of latent states selected from the encoded dataset $\mathcal Z_{\mathcal D}$, while $\mathcal U$ denotes the learned action codebook.
Let $q_t=\phi_Z(z_t)$ denote the vector-quantized state of $z_t$.
The vector-quantized action $u_t$ represents a transition between vector-quantized states, whereas $a_t$ is the physical action produced by the controller $\pi$.
% The state and action codebooks are $\mathcal{Q}$ and $\mathcal{U}$, and vector-quantized states are selected from the encoded dataset $\mathcal{Z}_{\mathcal D}$. 
% Let $q_t$ denote the vector-quantized state assigned to $z_t$.
% The vector-quantized action $u_t$ describes a transition between vector-quantized states; it is distinct from the robot action $a_t$ produced by $\pi$.
% The state quantization procedure is detailed in Sec.~\ref{sec:fpt} \ben{dont forward reference}.
% \paragraph{Quantized inverse dynamics model}
% Let $t' = \min\{s > t : q_s \neq q_t\}$ be the next change of vector-quantized state along the trajectory.
% We train on transitions $(q_t,q_{t'})$.
Let $t'=\min\{s>t:q_s\neq q_t\}$ denote the first subsequent time step at which the vector-quantized state changes.
We train on the resulting transitions $(q_t,q_{t'})$.
Given such a pair, $E_U$ produces a continuous latent action $\tilde{u}_t=E_U(q_t, q_{t'})$, which $\phi_U$ assigns to its nearest entry in the action codebook, $u_t =\phi_U(\tilde{u}_t) =  \arg\min_{u \in \mathcal{U}} \|\tilde{u}_t - u\|_2^2 $.
We train the action codebook with the standard vector quantization objective~\cite{vqvae},
\begin{equation}
  \mathcal{L}_{\mathrm{VQ}}
  = \|\mathrm{sg}(\tilde{u}_t) - u_t\|_2^2
  + \beta \|\tilde{u}_t - \mathrm{sg}(u_t)\|_2^2 ,
  \label{eq:vq_loss}
\end{equation}
where $\mathrm{sg}$ stops gradients and $\beta$ weights the commitment loss.
% \paragraph{Quantized latent predictor}
% \ben{I dont like the (1,2,3) then (a,b) listings}
The discrete latent predictor $P_{Q}$ outputs a categorical distribution over the vector-quantized states in $\mathcal{Q}$, conditioned on the current vector-quantized state and vector-quantized action, $P_{Q}(q_{t'} \mid q_t, u_t)$.
It is trained with the cross-entropy loss $\mathcal{L}_{\mathrm{transition}}
  = -\log P_{Q}\big(q_{t'} \mid q_t, u_t\big)$, and the joint training objective is $\mathcal{L}_{\mathrm{transition}} + \mathcal{L}_{\mathrm{VQ}}$.

\section{Experiments}

% We evaluate Q-WM on simulation benchmarks spanning manipulation, locomotion, and navigation, as well as a physical Tower of Hanoi task.
% These experiments examine long-horizon planning and execution, together with the ability to incorporate additional sensory inputs for control.
% An ablation on simulated Tower of Hanoi examines why state quantization should use first-passage time rather than Euclidean distance.
% We vary codebook size and training data to study how this choice affects planning and execution.

% \ben{consider flipping order of simulation and hardware experiments. right now the figure from hardware is too far from hardware section. usual order is simulation then hardware}

% We evaluate Q-WM through extensive experiments on a real-world Tower of Hanoi puzzle and simulation benchmarks spanning manipulation, and locomotion. 
% These experiments examine long-horizon planning and execution, as well as control with additional sensory inputs. 
% Ablations in simulated Tower of Hanoi compare first-passage-time and Euclidean quantization while varying codebook size and training data.
We evaluate VQ-WM through real-world and simulated experiments to investigate three questions:
\begin{description}
  \IEEEsetlabelwidth{Q3:}
  \item[\textbf{Q1:}] \textbf{Compositionality.} Can VQ-WM compose local transitions into long-horizon solutions beyond demonstrated trajectory lengths?
  \item[\textbf{Q2:}] \textbf{Long-horizon execution.} Can discrete planning with continuous feedback control sustain performance across long-horizon manipulation tasks?
  \item[\textbf{Q3:}] \textbf{Controller-aware abstraction.} Does first-passage-time-based vector quantization provide a more compact and effective planning abstraction than Euclidean vector quantization?
\end{description}
\noindent We investigate Q1 and Q2 through real-world Tower of Hanoi experiments and simulated manipulation benchmarks, and Q3 through controlled vector quantization ablations in simulated Tower of Hanoi. 
We additionally examine the flexibility of separating planning and control by incorporating proprioceptive observations in the Walker locomotion task.

\subsection{Simulation Benchmarks}
% \subsubsection{Simulation Benchmarks}
% \leavevmode\par
% \textit{How does our model architecture perform compare to baselines?}
% \paragraph{Simulation Benchmarks}
\textbf{Long-horizon manipulation.}
% \ben{Where do benchmarks come from?}
We evaluate manipulation in the Cube and Scene environments from Stable-WorldModel~\cite{stablewm}, using task-specific evaluation protocols.
Cube-Easy follows LeWM's protocol~\cite{lewm}, with a goal 25 environment steps ahead and an execution budget of 50 steps.
Cube-Hard follows H-JEPA's protocol~\cite{hjepa}, using 50 held-out, 20-step windows centered on a grasp and an execution budget of 30 steps to evaluate complete pick-and-place behavior.
For Scene, we define an evaluation with initial and goal images sampled 50 environment steps apart from expert trajectories and an execution budget of 100 steps.
Reaching the goal typically requires sequential subtasks involving the cube, buttons, drawer, or window, with success requiring all objects to simultaneously match their goal states.

VQ-WM achieves the highest success on all three tasks (Table~I), and its margin over the strongest baseline grows with the task horizon: 8.0 percentage points on Cube-Easy, 13.3 on Cube-Hard, and 28.0 on Scene.
We attribute these gains to two mechanisms.
First, VQ-WM decouples representation from planning geometry.
Temporal straightening (TS)~\cite{temporal} couples the two by reshaping the latent space so that Euclidean distance becomes a useful planning metric, whereas VQ-WM keeps a frozen encoder and learns the geometry separately through first-passage time under the controller.
VQ-WM improves over TS by 16.7 and 28.0 percentage points on Cube-Easy and Scene, and reaches 73.3\% on Cube-Hard, where TS largely fails at 10.7\%.
Second, VQ-WM replaces long-horizon action optimization with discrete search over controller-reachable vector-quantized states.
Cube-Easy goals often involve little change in cube position, so short-horizon planners already perform well, whereas Scene requires sequential subtasks over which errors in continuous rollouts accumulate.
Compared with the predicted subgoals of H-JEPA~\cite{hjepa}, which the controller may fail to reach, planning over reachable vector-quantized states is consistent with the 13.3-point gain on Cube-Hard over this strongest hierarchical baseline.

% \ben{75\% of this paragraph is restating infomration from the table. be more specific about why you think we have benefits, 
% e.g. Whereas the temporal straightening (TS) baseline couples latent representation and geometry by learning a latent space where Euclidean distance is a good metric (be more specific), our method decouples representation and geometry by learning a latent representation and then learning an appropriate geometry through the mean-first passage time distance. The 14\% and 28\% success rate improvement of QWM (Ours) compared to temporal straightening (TS) on Cube-Easy and Scene, and the ability to solve Cube-Hard, shows our approach is more robust for these problems.
% }

\begin{table}[!htb]
\centering
\caption{Long-horizon Manipulation success rates.}
\label{tab:simulation_baselines}
\footnotesize
\setlength{\tabcolsep}{3.0pt}
\renewcommand{\arraystretch}{1.15}
\begin{tabular*}{\linewidth}{@{\extracolsep{\fill}}lccc@{}}
\toprule
\textbf{Method}
& \textbf{Cube-Easy}
& \textbf{Cube-Hard}
& \textbf{Scene} \\
\midrule
LeWM~\cite{lewm} & $74.0\pm2.8$ & $32.0\pm6.1$ & $42.7\pm1.9$ \\
TS~\cite{temporal} & $68.0\pm2.8$ & $10.7\pm0.9$ & $47.3\pm2.5$ \\
% DINO-WM~\cite{dino-wm} & \textcolor{red}{--} & \textcolor{red}{--} & \textcolor{red}{--} \\
PLDM~\cite{pldm} & $62.0\pm4.3$ & $11.3\pm3.8$ & $44.7\pm4.1$ \\
H-JEPA~\cite{hjepa} & $76.7\pm2.5$ & $60.0\pm5.0$ &  $40.7\pm2.5$\\
\midrule
\textbf{Ours} & $\mathbf{84.7\pm2.5}$ & $\mathbf{73.3\pm0.9}$ & $\mathbf{75.3\pm5.0}$ \\
\bottomrule
\end{tabular*}
\end{table}

% \textbf{Navigation.}
% Q-WM and Temporal Straightening (TS)~\cite{temporal} both shape latent planning geometry, through first-passage-time-based state abstraction and trajectory straightening, respectively.
% On the complex manipulation task Scene, however, TS achieves only $47.3\%$ success compared with Q-WM's $75.3\%$, suggesting that trajectory straightening alone may be insufficient for reliable long-horizon manipulation.
% On PointMaze-Medium (Fig.~\ref{fig:navigation_horizons}), increasing the initial--goal separation from 25 to 75 environment steps leaves Q-WM's success nearly unchanged ($86.7\%$ to $85.3\%$), while TS drops from $100.0\%$ to $81.3\%$.
% Q-WM also achieves $100.0\%$ at both horizons on PointMaze-UMaze. \ben{these results are the least exciting of the paper for me. Are they worth including?}

% \begin{figure}[!htb]
% \centering
% \includegraphics[width=\linewidth]{figs/navigation_horizons.pdf}
% \caption{Navigation success on PointMaze-UMaze (left) and PointMaze-Medium (right) as the initial--goal separation increases from 25 to 75 environment steps. \ben{this plot is not well made.}}
% \label{fig:navigation_horizons}
% \end{figure}

\textbf{Locomotion.}
We evaluate goal-conditioned motion control in the DeepMind Control Walker environment from StableWorldModel~\cite{stablewm} to examine the advantage of separating planning from control.
Goal images are sampled 50 steps ahead on expert walking trajectories, and success requires the torso height, torso pitch, and all six leg joint angles to be within 0.3 of their goal values.
With visual inputs only, both DINO-WM~\cite{dino-wm} and VQ-WM perform poorly, reaching 12.7\% and 22.0\%, since such poses depend on precise joint coordination that images convey poorly.
After adding proprioception, DINO-WM remains at 12.0\%, whereas VQ-WM rises to 97.3\% (Table~\ref{tab:walker}).
The difference arises because VQ-WM handles control separately and its vector quantization is controller-aware: the additional input improves the controller, and the vector-quantized states adapt to what it can now reach without any change to the planner.
The same property allows VQ-WM to be readily extended to other sensing modalities at the control level.
% \ben{this demonstration is not related to our research, can you make the connection?}
% \xuezhou{This experiment aims to address the advantage of separating planning and controlling, and demonstrating the potential in extending to multi-modality controller inputs.}

% \begin{figure}[!htb]
% \centering
% \includegraphics[width=\linewidth]{figs/walker_modalities.pdf}
% \caption{Walker success with image-only and proprioceptive inputs. \ben{too much white space lets do a table for this}}
% \label{fig:walker_modalities}
% \end{figure}

% \begin{figure}[!htb]
% \centering
% \includegraphics[width=\linewidth]{figs/Hanoi_four_frames.pdf}
% \caption{Real-world execution of Q-WM on the four-disk Tower of Hanoi puzzle.}
% \label{fig:real_hanoi}
% \end{figure}

% \begin{table}[t]
%   \centering
%   \caption{Success rate (\%) on Walker with different observation inputs.}
%   \label{tab:walker}
%   \begin{tabular}{lcc}
%     \toprule
%     Method  & Images & Images + Proprio. \\
%     \midrule
%     DINO-WM & 12.7 $\pm$ 3.4 & 12.0 $\pm$ 4.9 \\
%     \textbf{Ours}    & \textbf{22.0$\pm$5.9} & \textbf{97.3$\pm$2.5} \\
%     \bottomrule
%   \end{tabular}
% \end{table}

\begin{table}[t]
  \centering
  \normalsize
  \caption{Success rate (\%) on Walker with different observation inputs.}
  \label{tab:walker}
  \begin{tabular*}{\columnwidth}{@{\extracolsep{\fill}}lcc}
    \toprule
    Method  & Images & Images + Proprio. \\
    \midrule
    DINO-WM & 12.7 $\pm$ 3.4 & 12.0 $\pm$ 4.9 \\
    \textbf{Ours}    & \textbf{22.0$\pm$5.9} & \textbf{97.3$\pm$2.5} \\
    \bottomrule
  \end{tabular*}
\end{table}

\subsection{Real-World Experiments}
% \cpw{Need to update this section}
% We evaluate VQ-WM on a physical four-disk Tower of Hanoi puzzle using a WidowX AI robotic arm~\cite{trossen_widowx_ai}, as illustrated in Fig.~\ref{fig:hanoi_real}.
% To evaluate long-horizon execution and the composition of local experience, we test VQ-WM on a physical four-disk Tower of Hanoi puzzle using a WidowX AI robotic arm~\cite{trossen_widowx_ai}, as illustrated in Fig.~\ref{fig:hanoi_real}.
\begin{figure}[!htb]
\centering
\includegraphics[width=\linewidth]{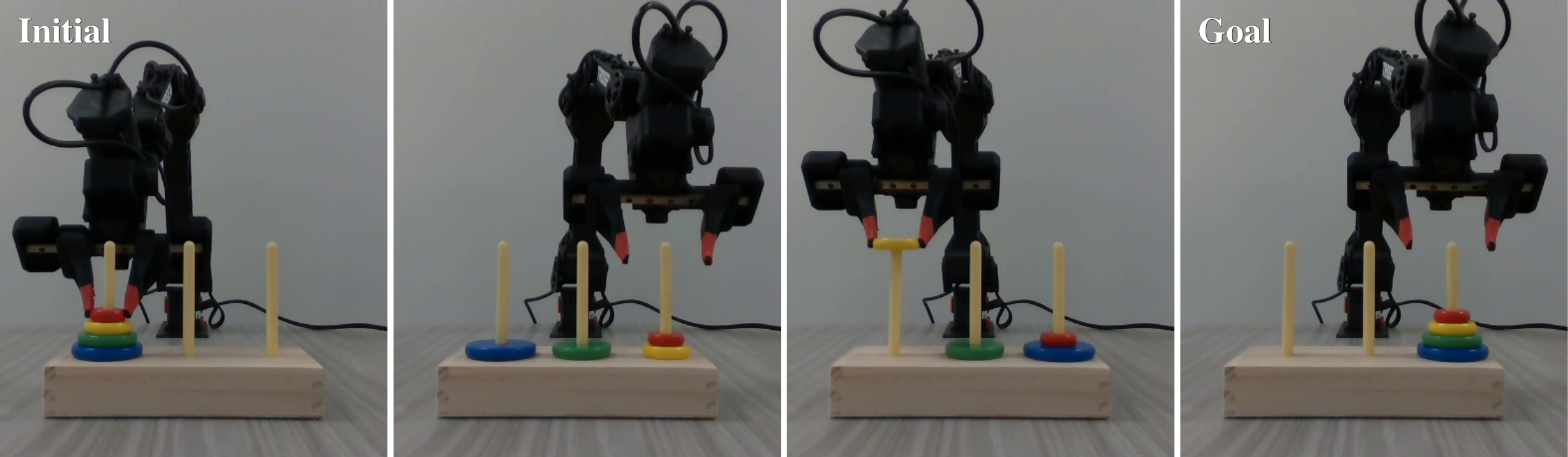}
\caption{Real-world execution of VQ-WM on the four-disk Tower of Hanoi puzzle.}
\label{fig:real_hanoi}
\end{figure}

To evaluate VQ-WM’s ability to compose and execute long-horizon plans from local behaviors, we test it on a real-world four-disk Tower of Hanoi puzzle using a WidowX AI robotic arm~\cite{trossen_widowx_ai}, as shown in Fig.~\ref{fig:real_hanoi}.
The goal is to transfer the entire stack to another rod, moving one disk at a time without placing a larger disk on a smaller one.
Solving the full puzzle optimally requires 15 puzzle moves (H15), combining planning over minutes of execution with precise grasping, alignment, and disk insertion.
We also evaluate single-move tasks (H1) to compare performance across short and long task horizons.

\textbf{Data Collection.} Tower of Hanoi play data was collected autonomously using a WidowX AI arm at 30\,Hz. 
A symbolic planner generated randomized legal move sequences using board states detected by the video object segmentation model Cutie~\cite{cutie} at episode boundaries. 
Each disk transfer was executed using a predefined Cartesian pick-and-place motion.

\begin{figure*}[t]
    \centering
    \includegraphics[width=\linewidth]{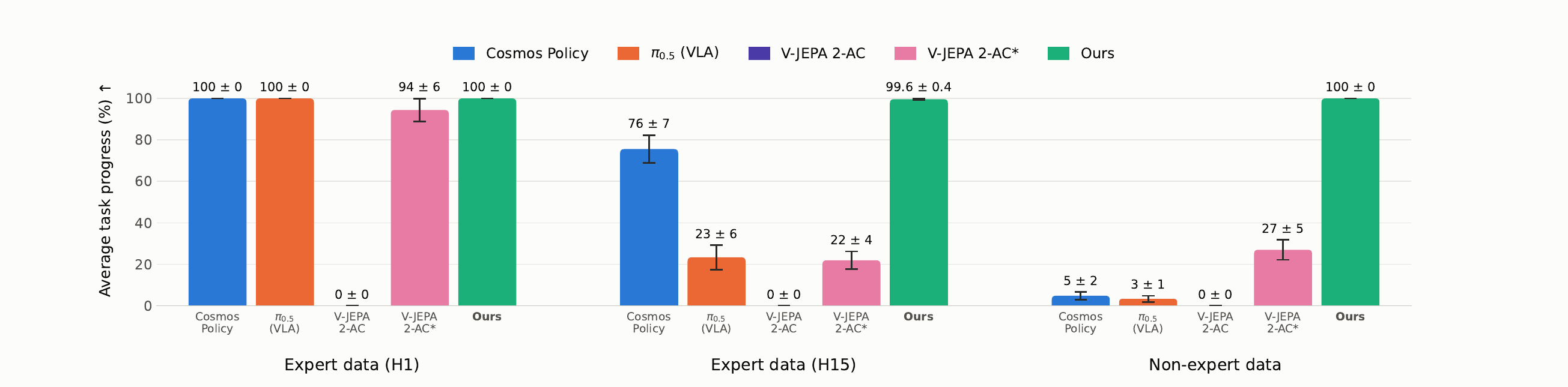}
    \caption{\textbf{Real-world results.} VLA and WAM baselines without runtime search fail to extrapolate from short play subtasks to full-puzzle solutions, whereas VQ-WM composes local behaviors into complete plans and executes them reliably. V-JEPA 2-AC struggles even on H1, suggesting a limited effective planning horizon. V-JEPA 2-AC* receives manually specified subgoals that supply the solution sequence, leaving the model to perform local planning and control.}
    % ben{VLAs and policies that do not use search at runtime cannot extrapolate beyond the dataset} }
    \label{fig:hanoi_real}
\end{figure*}

\textbf{Evaluation Protocol.} We compare VQ-WM against the world action model (WAM) Cosmos Policy~\cite{cosmos-policy}, the vision-language-action model (VLA) $\pi_{0.5}$~\cite{pi_05}, and the world foundation model V-JEPA 2-AC~\cite{vjepa2} under expert training on H1 and H15 and non-expert training on H15. 
The expert dataset contains complete solution demonstrations for six initial–goal configuration pairs. 
The VLA and WAM baselines are trained jointly on these six tasks. 
In the non-expert dataset setting, the VLA and WAM baselines are trained on play-data subtasks spanning at most five puzzle moves and evaluated on full H15 tasks, testing whether they can compose shorter subtasks into long-horizon solutions.
To reduce unsupported predictions during planning, we augment VQ-WM’s discrete latent predictor with an additional rejection class for unseen transitions and exclude predictions in this class from graph search.
For V-JEPA 2-AC, we restrict the action sampler to actions observed in the dataset.
Following~\cite{vjepa2}, we also evaluate V-JEPA 2-AC* with six manually specified subgoal images per puzzle move.
These subgoals provide the high-level solution structure, leaving the model to perform local planning and control between successive targets.
We evaluate each method on the same six test cases with three trials per case, using a single NVIDIA RTX 5080 GPU for inference.
We present the results in Fig.~\ref{fig:hanoi_real}.
The Tower of Hanoi puzzle challenges both planning and control: solving it requires planning far ahead while repeatedly executing precise transfers of small disks between pegs.

\textbf{Expert data setting.} Cosmos Policy and $\pi_{0.5}$ both achieve 100\% task progress on H1, but drop to 76\% and 23\%, respectively, on H15. 
This decline suggests that execution errors compound across successive disk transfers, reducing reliability over the full puzzle.
V-JEPA 2-AC uses the cross-entropy method (CEM)~\cite{CEM} to optimize continuous action sequences over a fixed horizon, executing only the first action before replanning, as in receding-horizon control. 
Extending its lookahead requires searching over longer action sequences, whose predicted outcomes become less reliable as prediction errors accumulate across autoregressive latent rollouts.
Frequent replanning alone cannot ensure progress when the benefits of necessary intermediate actions lie beyond this horizon. 
This is particularly relevant to Tower of Hanoi, where intermediate puzzle moves and physical grasping and lifting motions may not immediately bring the observed state closer to the final goal. 
V-JEPA 2-AC achieves 0\% task progress on both H1 and H15, suggesting that even a single puzzle move lies beyond its effective planning horizon.
By providing manually selected dataset images corresponding to the descend, grasp, lift, transit, insert, and release stages of each disk transfer as sequential subgoals, we raise V-JEPA 2-AC*’s task progress to $94\% \pm 6\%$ on H1.
% By providing subgoal images manually selected from the dataset as sequential planning targets, V-JEPA 2-AC* raises its task progress to $94\% \pm 6\%$.
% By providing manual subgoals, V-JEPA 2-AC* raises its task progress to $94\% \pm 6\%$.
% \ben{how are manual subgoals provided?}
The gap between the two settings points to a limitation: the model can reach a nearby target when one is provided, but cannot discover the intermediate moves on its own.
Even with manually provided subgoals, V-JEPA 2-AC* struggles to execute full H15 solutions reliably.
VQ-WM instead searches for the shortest path through controller-aware latent abstractions, allowing the planner to reason far into the future without optimizing the full sequence of low-level actions. 
This also makes our planner efficient, allowing us to use an infinite replanning interval and compute the complete solution in a single planning call.
A continuous controller then executes the plan by sequentially reaching the targets using visual feedback to correct local deviations and mitigate compounding execution errors.
VQ-WM achieves 99.6\% task progress on H15, demonstrating reliable execution over the full 15-move horizon.

% \ben{be more specific about the mechanism that differentiates V-JEPA-2AC and Ours here, and how that mechanism buys us long horizon stability}.

\textbf{Non-Expert play data setting.} Cosmos Policy and $\pi_{0.5}$ achieve only 5\% and 3\% task progress, respectively, indicating that these behavior-cloning policies struggle to compose shorter subtasks into complete H15 solutions. 
Although V-JEPA 2-AC can plan beyond demonstrated trajectories, it still struggles with long-horizon planning and execution, as discussed above. 
Providing manual subgoals improves task progress to $27\%$ for V-JEPA 2-AC*, though reliable execution over the full H15 horizon remains challenging.
VQ-WM achieves 100\% task progress, demonstrating its ability to compose observed transitions into long-horizon solutions and reliably execute the resulting plans.

\subsection{Ablation on State Vector Quantization}
% \xuezhou{Since we mixed up L2 ($\epsilon$-neighborhood) and FPT, FPT will reduce to L2-distance when the codebook size getting bigger, which explains the overlap of right side of the curve.}
% \xuezhou{I think the good way to illustrate our contribution to use FPT is:
% i) Conceptually more explainable to align planning and control;
% ii) Better generalization ability on tasks that are more complex or with limited data.}

\label{abl:state_quantization}
% \subsubsection{Simulated Tower of Hanoi}
% \label{sec:simulated_hanoi}
% We conduct the state quantization ablation in Sec.~\ref{abl:state_quantization} in MuJoCo~\cite{mujoco} to enable extensive trials for statistically meaningful comparisons.
% The simulation closely follows our real-world setup, using the same WidowX AI arm model and four-disk, three-peg configuration.
% The camera viewpoint, arm placement, and puzzle dimensions approximate the physical setup without exact geometric calibration.
% Observations are $224\times224$ side-view images.
% The offline dataset consists of 104 walk trajectories covering all legal disk-move transitions.

% For each of three seeds, we randomly sample 50 initial--goal configuration pairs, covering shortest-path lengths from 1 to 15 disk moves. 
% We report \emph{optimal-path success}: the robot must reach the correct goal configuration using the minimum number of disk moves. 

% We examine how state quantization affect the compatibility between discrete planning and feedback control.
% These ablations evaluate the trade-off between executable state abstraction and reliable transition learning, as well as the effects of predicted shortcuts on execution.
% The Tower of Hanoi ablations use the simulated setup described in Sec.~\ref{sec:simulated_hanoi}.
To enable extensive trials for statistically meaningful comparisons, we conduct this ablation study on a simulated version of Tower of Hanoi in MuJoCo~\cite{mujoco} with the same robotic arm and puzzle setup as our real-world experiment.
% The simulation closely follows our real-world setup, using the same WidowX AI arm model and four-disk, three-peg configuration.
% The camera viewpoint, arm placement, and puzzle dimensions approximate the physical setup without exact geometric calibration.
% Observations are $224\times224$ side-view images.
The dataset contains 104 walk trajectories covering all legal disk-move transitions.
We compare first-passage time with squared Euclidean distance for codebook construction and state assignment, keeping the encoder, controller, and training procedure unchanged. Each point is evaluated over three seeds with 50 trajectories per seed, using randomly sampled initial and goal configurations spanning shortest-path lengths of 1–15 disk moves.

\begin{figure}[!htb]
    \centering
    \includegraphics[width=\linewidth]{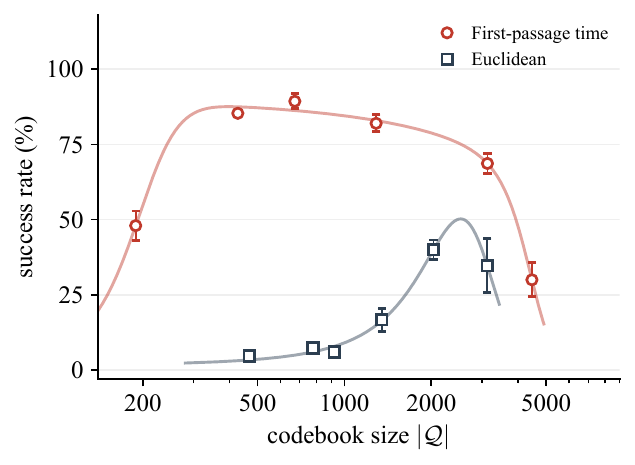}
    \caption{State vector quantization with different codebook sizes.}
    \label{fig:state_quantization_sweep}
\end{figure}

Fig.~\ref{fig:state_quantization_sweep} shows that first-passage time achieves approximately 90\% peak success, compared with 40\% for Euclidean distance, an improvement of about 50 percentage points. It also sustains high success across a broad range of codebook sizes, roughly 250–2000, spanning an eightfold change. Performance decreases at both extremes: overly small codebooks merge distinct puzzle configurations, producing unexecutable plans, while overly large codebooks leave fewer observed examples per transition, reducing the accuracy of the discrete latent predictor.

These results highlight the benefits of using first-passage time to align state abstraction and feedback control with the controller’s capabilities. First, Euclidean distance ignores controller reachability, so consecutive planned targets may be unreachable under the controller. Second, during control loop, the Euclidean variant declares a target reached when $\|z_t-q_i\|_2\leq\epsilon$. Since latent proximity does not always indicate physical arrival, the target may switch too early or fail to switch after arrival. Instead, in the first-passage-time variant, the number of controller calls required to reach each target is estimated using the continuous latent predictor.

% \ben{[Done] can you try a non towers of hanoi task to get more compelling justification of first passage time?}
% \xuezhou{Changes: Unify arrival condition, L2 use $|z-z'|<\epsilon$ while FPT uses continuous latent predictor estimation, and use the same total steps budget to test SR (if didn't success after certain steps, will be marked fail.}

% We examine the importance of first-passage time for state vector quantization by comparing it with squared Euclidean distance on simulated Tower of Hanoi.
% We sweep the codebook size $K_z=|\mathcal{E}_z|$ and training data size to characterize this comparison across different abstraction granularities and data budgets.
% The two variants use the following distances:

\section{Limitations and Future Work}
First, the computational efficiency of its discrete search makes frequent MPC-style replanning practical. Our experiments use a single plan per episode to evaluate the reliability of our feedback motion planning framework, leaving the benefits of replanning unexplored. 
Future work can investigate how replanning frequency affects computational cost and recovery from deviations that require revising the planned sequence.
Second, VQ-WM's modular design offers a direct path toward multi-task settings. 
A pretrained vision-language model can provide instruction-conditioned representations, while a multi-task policy can serve as the controller. Controller-aware vector quantization can then adapt the planning space to the selected controller with planning process unchanged. 
Evaluating these combinations across tasks is a promising direction for extending VQ-WM's applicability.
% \ben{this limitation section is too negative, make it more about future work. "successful trajectories" is wrong word for this because our whole point is that world model relaxes need for expert data by composing information.} \ben{include something about MPC style, we can actually run really fast, we just did open loop here to show stability of feedback motion planning}. 

\section{Conclusion}
We presented VQ-WM, a latent feedback motion planning framework that integrates discrete long-horizon search with a continuous controller for closed-loop execution.
First-passage time estimated under learned closed-loop dynamics connects these two roles: it defines controller-dependent coverage regions and uses them to construct vector-quantized planning states.
Discrete search uses the discrete latent predictor to plan over representative states, while feedback control executes the resulting targets.
Real-world Tower of Hanoi experiments demonstrate the effectiveness of this integration of planning and control, with VQ-WM outperforming state-of-the-art VLA, WAM, and world foundation model baselines.
Simulation benchmarks spanning manipulation, and locomotion further demonstrate its broader applicability. 
These results support controller-dependent state abstraction as a practical foundation for long-horizon planning and reliable closed-loop execution.

\bibliographystyle{IEEEtran}
\bibliography{references}

@article{lewm,
  title   = {{LeWorldModel}: Stable End-to-End Joint-Embedding Predictive Architecture from Pixels},
  author  = {Maes, Lucas and Le Lidec, Quentin and Scieur, Damien and LeCun, Yann and Balestriero, Randall},
  journal = {arXiv preprint arXiv:2603.19312},
  year    = {2026},
  url     = {https://arxiv.org/abs/2603.19312}
}

@misc{dino-wm,
      title={DINO-WM: World Models on Pre-trained Visual Features enable Zero-shot Planning}, 
      author={Gaoyue Zhou and Hengkai Pan and Yann LeCun and Lerrel Pinto},
      year={2024},
      eprint={2411.04983},
      archivePrefix={arXiv},
      primaryClass={cs.RO},
      url={https://arxiv.org/abs/2411.04983}, 
}

@INPROCEEDINGS{ lqrtrees,
    AUTHOR    = {R. Tedrake},
    TITLE     = {{LQR}-trees: Feedback motion planning on sparse randomized trees},
    BOOKTITLE = {Proceedings of Robotics: Science and Systems},
    YEAR      = {2009},
    ADDRESS   = {Seattle, USA},
    MONTH     = {June},
    DOI       = {10.15607/RSS.2009.V.003} 
}

@inproceedings{vqvae,
  title     = {Neural Discrete Representation Learning},
  author    = {van den Oord, Aaron and Vinyals, Oriol and Kavukcuoglu, Koray},
  booktitle = {Advances in Neural Information Processing Systems},
  volume    = {30},
  year      = {2017}
}

@article{hwm,
  title={Hierarchical planning with latent world models},
  author={Zhang, Wancong and Terver, Basile and Zholus, Artem and Chitnis, Soham and Sutaria, Harsh and Assran, Mido and Balestriero, Randall and Bar, Amir and Bardes, Adrien and LeCun, Yann and others},
  journal={arXiv preprint arXiv:2604.03208},
  year={2026}
}

@article{stablewm,
  title={stable-worldmodel: A Platform for Reproducible World Modeling Research and Evaluation},
  author={Maes, Lucas and Lidec, Quentin Le and Facury, Luiz and Massaudi, Nassim and Chaurasia, Ayush and Capuano, Francesco and Gao, Richard and Gillin, Taj and Haramati, Dan and Scieur, Damien and others},
  journal={arXiv preprint arXiv:2605.21800},
  year={2026}
}

@article{vjepa2,
  title   = {{V-JEPA 2}: Self-Supervised Video Models Enable Understanding, Prediction and Planning},
  author  = {Assran, Mido and Bardes, Adrien and Fan, David and Garrido, Quentin and Howes, Russell and Komeili, Mojtaba and others},
  journal = {arXiv preprint arXiv:2506.09985},
  year    = {2025},
  url     = {https://arxiv.org/abs/2506.09985}
}

@article{director,
  title={Deep hierarchical planning from pixels},
  author={Hafner, Danijar and Lee, Kuang-Huei and Fischer, Ian and Abbeel, Pieter},
  journal={Advances in Neural Information Processing Systems},
  volume={35},
  pages={26091--26104},
  year={2022}
}

@inproceedings{pldm,
  title     = {Learning from Reward-Free Offline Data: A Case for Planning with Latent Dynamics Models},
  author    = {Sobal, Uladzislau and Zhang, Wancong and Cho, Kyunghyun and Balestriero, Randall and Rudner, Tim G. J. and LeCun, Yann},
  booktitle = {Advances in Neural Information Processing Systems},
  volume    = {38},
  year      = {2025},
  doi       = {10.52202/085713-1465}
}

@article{temporal,
  title={Temporal straightening for latent planning},
  author={Wang, Ying and Bounou, Oumayma and Zhou, Gaoyue and Balestriero, Randall and Rudner, Tim GJ and LeCun, Yann and Ren, Mengye},
  journal={arXiv preprint arXiv:2603.12231},
  year={2026}
}

@inproceedings{tdmpc,
  title     = {Temporal Difference Learning for Model Predictive Control},
  author    = {Hansen, Nicklas and Wang, Xiaolong and Su, Hao},
  booktitle = {Proceedings of the 39th International Conference on Machine Learning},
  series    = {Proceedings of Machine Learning Research},
  volume    = {162},
  year      = {2022},
  url       = {https://proceedings.mlr.press/v162/hansen22a.html}
}

@inproceedings{tdmpc2,
  title     = {{TD-MPC2}: Scalable, Robust World Models for Continuous Control},
  author    = {Hansen, Nicklas and Su, Hao and Wang, Xiaolong},
  booktitle = {International Conference on Learning Representations},
  year      = {2024},
  url       = {https://arxiv.org/abs/2310.16828}
}

@article{cosmos-policy,
  title={Cosmos policy: Fine-tuning video models for visuomotor control and planning},
  author={Kim, Moo Jin and Gao, Yihuai and Lin, Tsung-Yi and Lin, Yen-Chen and Ge, Yunhao and Lam, Grace and Liang, Percy and Song, Shuran and Liu, Ming-Yu and Finn, Chelsea and others},
  journal={arXiv preprint arXiv:2601.16163},
  year={2026}
}

@INPROCEEDINGS{mujoco,
  author={Todorov, Emanuel and Erez, Tom and Tassa, Yuval},
  booktitle={2012 IEEE/RSJ International Conference on Intelligent Robots and Systems}, 
  title={MuJoCo: A physics engine for model-based control}, 
  year={2012},
  volume={},
  number={},
  pages={5026-5033},
  doi={10.1109/IROS.2012.6386109}}

@misc{trossen_widowx_ai,
  author       = {{Trossen Robotics}},
  title        = {{WidowX AI} Robotic Arm},
  howpublished = {\url{https://www.trossenrobotics.com/widowx-ai}},
  note         = {Accessed: October 6, 2026}
}

@inproceedings{cutie,
  title     = {Putting the Object Back into Video Object Segmentation},
  author    = {Cheng, Ho Kei and Oh, Seoung Wug and Price, Brian and Lee, Joon-Young and Schwing, Alexander},
  booktitle = {Proceedings of the IEEE/CVF Conference on Computer Vision and Pattern Recognition (CVPR)},
  year      = {2024}
}

@InProceedings{pi_05,
  title = 	 {$\pi_{0.5}$: a Vision-Language-Action Model with Open-World Generalization},
  author = {Black, Kevin and Brown, Noah and Darpinian, James and Dhabalia, Karan and Driess, Danny and Esmail, Adnan and others},
  booktitle = 	 {Proceedings of The 9th Conference on Robot Learning},
  pages = 	 {17--40},
  year = 	 {2025},
  editor = 	 {Lim, Joseph and Song, Shuran and Park, Hae-Won},
  volume = 	 {305},
  series = 	 {Proceedings of Machine Learning Research},
  month = 	 {27--30 Sep},
  publisher =    {PMLR},
  url = 	 {https://proceedings.mlr.press/v305/black25a.html}
}

@inproceedings{planet,
  title     = {Learning Latent Dynamics for Planning from Pixels},
  author    = {Hafner, Danijar and Lillicrap, Timothy and Fischer, Ian and Villegas, Ruben and Ha, David and Lee, Honglak and Davidson, James},
  booktitle = {Proceedings of the 36th International Conference on Machine Learning},
  series    = {Proceedings of Machine Learning Research},
  volume    = {97},
  pages     = {2555--2565},
  year      = {2019},
  url       = {https://proceedings.mlr.press/v97/hafner19a.html}
}

@inproceedings{dreamer,
  title     = {Dream to Control: Learning Behaviors by Latent Imagination},
  author    = {Hafner, Danijar and Lillicrap, Timothy and Ba, Jimmy and Norouzi, Mohammad},
  booktitle = {International Conference on Learning Representations},
  year      = {2020},
  url       = {https://openreview.net/forum?id=S1lOTC4tDS}
}

@inproceedings{e2c,
  title     = {Embed to Control: A Locally Linear Latent Dynamics Model for Control from Raw Images},
  author    = {Watter, Manuel and Springenberg, Jost Tobias and Boedecker, Joschka and Riedmiller, Martin},
  booktitle = {Advances in Neural Information Processing Systems},
  volume    = {28},
  year      = {2015},
  url       = {https://arxiv.org/abs/1506.07365}
}

@article{visualforesight,
  title   = {Visual Foresight: Model-Based Deep Reinforcement Learning for Vision-Based Robotic Control},
  author  = {Ebert, Frederik and Finn, Chelsea and Dasari, Sudeep and Xie, Annie and Lee, Alex and Levine, Sergey},
  journal = {arXiv preprint arXiv:1812.00568},
  year    = {2018},
  url     = {https://arxiv.org/abs/1812.00568}
}

@inproceedings{upn,
  title     = {Universal Planning Networks: Learning Generalizable Representations for Visuomotor Control},
  author    = {Srinivas, Aravind and Jabri, Allan and Abbeel, Pieter and Levine, Sergey and Finn, Chelsea},
  booktitle = {Proceedings of the 35th International Conference on Machine Learning},
  series    = {Proceedings of Machine Learning Research},
  volume    = {80},
  pages     = {4732--4741},
  year      = {2018},
  url       = {https://proceedings.mlr.press/v80/srinivas18b.html}
}

@inproceedings{play_lmp,
  title     = {Learning Latent Plans from Play},
  author    = {Lynch, Corey and Khansari, Mohi and Xiao, Ted and Kumar, Vikash and Tompson, Jonathan and Levine, Sergey and Sermanet, Pierre},
  booktitle = {Proceedings of the Conference on Robot Learning},
  series    = {Proceedings of Machine Learning Research},
  volume    = {100},
  pages     = {1113--1132},
  year      = {2020},
  url       = {https://proceedings.mlr.press/v100/lynch20a.html}
}

@inproceedings{glamor,
  title     = {Planning from Pixels using Inverse Dynamics Models},
  author    = {Paster, Keiran and McIlraith, Sheila A. and Ba, Jimmy},
  booktitle = {International Conference on Learning Representations},
  year      = {2021},
  url       = {https://openreview.net/forum?id=V6BjBgku7Ro}
}

@inproceedings{sorb,
  title     = {Search on the Replay Buffer: Bridging Planning and Reinforcement Learning},
  author    = {Eysenbach, Ben and Salakhutdinov, Ruslan and Levine, Sergey},
  booktitle = {Advances in Neural Information Processing Systems},
  volume    = {32},
  year      = {2019},
  url       = {https://proceedings.neurips.cc/paper_files/paper/2019/hash/5c48ff18e0a47baaf81d8b8ea51eec92-Abstract.html}
}

@inproceedings{l3p,
  title     = {World Model as a Graph: Learning Latent Landmarks for Planning},
  author    = {Zhang, Lunjun and Yang, Ge and Stadie, Bradly C.},
  booktitle = {Proceedings of the 38th International Conference on Machine Learning},
  series    = {Proceedings of Machine Learning Research},
  volume    = {139},
  pages     = {12611--12620},
  year      = {2021},
  url       = {https://proceedings.mlr.press/v139/zhang21x.html}
}

@article{sequential_composition,
  title   = {Sequential Composition of Dynamically Dexterous Robot Behaviors},
  author  = {Burridge, Robert R. and Rizzi, Alfred A. and Koditschek, Daniel E.},
  journal = {The International Journal of Robotics Research},
  volume  = {18},
  number  = {6},
  pages   = {534--555},
  year    = {1999},
  doi     = {10.1177/02783649922066385}
}

@article{funnels,
  title   = {Funnel Libraries for Real-Time Robust Feedback Motion Planning},
  author  = {Majumdar, Anirudha and Tedrake, Russ},
  journal = {The International Journal of Robotics Research},
  volume  = {36},
  number  = {8},
  pages   = {947--982},
  year    = {2017},
  doi     = {10.1177/0278364917712421}
}

@article{hjepa,
  title   = {{H-JEPA}: End-to-End Learning of Hierarchical World Models for Visual Planning},
  author  = {Zhang, Wancong and Terver, Basile and Rabbat, Michael and LeCun, Yann and Balestriero, Randall},
  journal = {arXiv preprint arXiv:2610.06805},
  year    = {2026},
  url     = {https://arxiv.org/abs/2610.06805}
}

@article{CEM,
title = {Optimization of computer simulation models with rare events},
journal = {European Journal of Operational Research},
volume = {99},
number = {1},
pages = {89-112},
year = {1997},
issn = {0377-2217},
doi = {https://doi.org/10.1016/S0377-2217(96)00385-2},
url = {https://www.sciencedirect.com/science/article/pii/S0377221796003852},
author = {Reuven Y. Rubinstein}
}

@inproceedings{thick,
  title     = {Learning Hierarchical World Models with Adaptive Temporal Abstractions from Discrete Latent Dynamics},
  author    = {Gumbsch, Christian and Sajid, Noor and Martius, Georg and Butz, Martin V.},
  booktitle = {International Conference on Learning Representations (ICLR)},
  year      = {2024}
}

@article{hilewm,
  title   = {Mind the Gap: Promises and Pitfalls of Hierarchical Planning in {LeWorldModel}},
  author  = {Caselli, Niccol{\`o} and Massafra, Francesco and Punzo, Samuele and Lo Sardo, Salvatore and Pantelidis, Ippokratis and Bhethanabhotla, Sathya Kamesh},
  journal = {arXiv preprint arXiv:2607.12547},
  year    = {2026}
}

@article{hydra,
  title={Hydra: A Navigation World Action Model with Discrete Latent Planning and Continuous Flow-Matching Execution},
  author={Nazeri, Mohammad and Card, Alexandyr and Huber, Samira and Pokhrel, Anuj and Wang, Yujun and Hammele, Ruben and Song, Daeun and Pirk, S{\"o}ren and Xiao, Xuesu},
  journal={arXiv preprint arXiv:2608.28995},
  year={2026}
}

\newpage

\vfill

\end{document}